\documentclass[final]{cvpr}

\usepackage{times}
\usepackage{epsfig}
\usepackage{graphicx}
\usepackage{amsmath}
\usepackage{amssymb}
\usepackage{multirow}
\usepackage{gensymb}
\usepackage{hhline}

\usepackage[pagebackref=true,breaklinks=true,colorlinks,bookmarks=false]{hyperref}

\def\cvprPaperID{8124} 
\def\confYear{CVPR 2021}

\begin{document}

\title{Zero-Shot Calibration of Fisheye Cameras}

\author{Jae-Yeong Lee\\
Electronics and Telecommunications Research Institute (ETRI)\\
Daejeon 34129, South Korea\\
{\tt\small jylee@etri.re.kr}
}

\maketitle

\begin{abstract}
In this paper, I present a novel zero-shot camera calibration method that estimates camera parameters with no calibration image. It is common sense that we need at least one or more pattern images for camera calibration. However, the proposed method estimates camera parameters from the horizontal and vertical field of view information of the camera without any image acquisition. The proposed method is particularly useful for wide-angle or fisheye cameras that have large image distortion. Image distortion is modeled in the way fisheye lenses are designed and estimated based on the square pixel assumption of the image sensors. The calibration accuracy of the proposed method is evaluated on eight different commercial cameras qualitatively and quantitatively, and compared with conventional calibration methods. The experimental results show that the calibration accuracy of the zero-shot method is comparable to conventional full calibration results. The method can be used as a practical alternative in real applications where individual calibration is difficult or impractical, and in most field applications where calibration accuracy is less critical. Moreover, the estimated camera parameters by the method can also be used to provide proper initialization of any existing calibration methods, making them to converge more stably and avoid local minima.
\end{abstract}

\section{Introduction}

Camera calibration is an essential and preliminary step in computer vision. In 3D vision applications, the camera parameters are used to interpret and recover imaging geometry of a camera from a 3D scene to a 2D image. In 2.5D vision applications like visual surveillance, camera calibration is required to normalize and eliminate the influence of the camera-specific parameters on the execution of the algorithm.

In general, camera calibration requires careful acquisition of dozens of pattern images and is usually performed in a desktop environment. However, in real applications, image acquisition for camera calibration is not an easy problem and can be challenging depending on the circumstances. To calibrate a camera, we usually need to take a number of images of special calibration patterns such as checkerboard, circles, or a known 3D object. The captured pattern images are then used to extract control points which constrain camera parameters and thus they have to be carefully taken such that the patterns do not deviate from the camera's field of view and as many as possible in various positions and angles~\cite{galeone_camera_2018}. This image acquisition process is time-consuming and often challenging especially when a camera is placed on unconstrained areas or fixed in high on the environment like most CCTV cameras. Recent intelligent vision systems usually utilize dozens or hundreds of cameras in one site and therefore camera calibration have become one of the most challenging process in such a field.

There have been lots of approaches to minimize the effort required to acquire images in the literature, namely few-shot calibration approaches. They can be classified into several groups. Most of them use orthogonal pairs of vanishing points in a natural scene or artificial objects~\cite{beardsley92, guillou00, kosecka_video_2002, lee2012camera}, or estimate planar homography and then extract camera parameters from the estimated homography~\cite{zhou_chuan_planar_2003, herrera_forget_2016, lv_methods_nodate, strobl_more_2011, zhang_homography-based_2008, zhengyou_zhang_flexible_1999}, or estimate camera parameters using a specially designed 2D or 3D object~\cite{grossmann_camera_2014, huang_research_2019}. These few-shot approaches are more preferred in practical applications because the conventional checkerboard-based calibration requires a lot of images taken with care. However, these simplified few-shot methods still require image acquisition together with a subsequent sophisticated image analysis process. What is worse is that they usually assume zero-distortion of the camera lens or simply ignore the image distortion, which is not the case for most recent wide-angle or fisheye cameras. In these days, more and more cameras are used in real life applications and they tends to use wide-angle lenses to secure a large field of views.

More recently lots of auto calibration approaches based on deep neural network~\cite{chao_self-supervised_2020, xiaoyu_blind_2019, liao_deep_2020, xue_learning_2019, xue_fisheye_2020, xiaoqing_fisheye_2018, zhao_simple_2020} have also been studied. However, these approaches mainly focus on the rectification of the image and their performance still rely on the acquired images.

In this paper, I adopt a completely different direction and suggest a novel calibration method that estimates camera parameters without any image acquisition, namely {\em zero-shot} camera calibration method. The method estimates camera parameters by utilizing only the camera specification information without any image acquisition. Since the proposed method does not require image acquisition, no subsequent image analysis nor feature detection which can be oftentimes erroneous is required. The experimental results on eight commercial cameras show that the proposed zero-shot calibration method is able to estimate their camera parameters effectively with an accuracy comparable to conventional image-based calibration methods.

The proposed method is particularly useful for wide-angle or fisheye cameras that have large image distortion. The image distortion of camera lens is modeled in the way fisheye lenses are designed and estimated based on the square pixel assumption of the image sensors. And the experimental result on real cameras shows that the zero-shot method yields better rectification results than a standard calibration method~\cite{zhang_flexible_2000}.

My work, to the best of my knowledge, is the first one to estimate camera parameters of fisheye cameras without image acquisition. The benefit of the proposed method is obvious and it can be used in various ways. It can be used as a practical alternative in real applications where individual calibration is difficult or impractical, and in most field applications where calibration accuracy is less critical. Moreover, the estimated camera parameters by the zero-shot method can also be used to provide proper initialization of any existing calibration methods, making them to converge more stably and avoid local minima. The effectiveness of this kind of initialization is also demonstrated in the experiment.

The paper is organized as follows. Section~\ref{sec:method} describes the proposed camera calibration method along with motivations and technical details. In Section~\ref{sec:experim}, experimental results on real cameras are presented. In the experiment, calibration accuracy is evaluated in terms of RMS pixel reprojection error and compared with conventional calibration methods. In Section~\ref{sec:singleshot}, a single-shot extension of the method is presented and evaluated. Finally, I conclude the paper with a brief discussion in Section~\ref{sec:discussion}.

\section{Zero-Shot Calibration Method}
\label{sec:method}

\subsection{Simplified camera model}

In this paper, a simplified pinhole camera projection model is employed where a camera is assumed to have zero skew and unit aspect ratio (square pixels). This assumption naturally holds for most modern cameras because pixel aspect ratios of photodiode cells of modern image sensors are typically one (square pixel) and also pixel skew is zero or safely negligible. The camera model is further assumed to have zero decentering distortion and the principal point to coincide with the image center.
\begin{align}
	&f = f_x = f_y\\
	&c_x = W/2\\
	&c_y = H/2
\end{align}
where $W$ is the image width and $H$ the image height.

The {\em square pixel} assumption on the image sensors of modern cameras is the most important premise that the proposed method is based on and motivated from. The physical focal length of a camera is the distance from the lens center to the image sensor when the image is best in focus. In computer vision, however, the focal length as a camera parameter is defined relatively in pixel unit as the ratio of physical focal length to the cell size of the image sensor. More specifically the ratio of the physical focal length to the image sensor cell size in the horizontal direction is defined by $f_x$ and the ratio to the image sensor cell size in the vertical direction is defined by $f_y$. Therefore, we can naturally assume that $f_x$ and $f_y$ have the same value for most modern cameras where image sensor cells are designed to have square size and arrayed in uniform grids.

\subsection{Inconsistency in focal length estimation}

The perspective projection of a pinhole camera is described by the following formula~\cite{kannala_generic_2006}
\begin{equation}
r = f \tan \theta
\label{eq:perspective}
\end{equation}
where $\theta$ is the angle between the principal axis and the incoming ray, $r$ is the distance between the image point and the principal point and $f$ is the focal length.

If a camera follows the perspective projection model, the horizontal and vertical focal length of the camera can be computed directly from its horizontal and vertical field of view angles as follows:
\begin{align}
{\tilde f}_x &= \frac{W/2}{\tan \left( \theta_h / 2 \right)}\label{eq:fx}\\
{\tilde f}_y &= \frac{H/2}{\tan \left( \theta_v / 2 \right)}\label{eq:fy}
\end{align}
where $\theta_h$ is horizontal field of view angle and $\theta_v$ vertical field of view angle of camera and $W$ is the image width and $H$ the image height. The field of view angles information of a camera are usually given in camera specification which is provided by the manufacturer.

The computed horizontal focal length~${\tilde f}_x$ and vertical focal length~${\tilde f}_y$ should have virtually the same or similar values if the camera follows the {\em square pixel} assumption. And this property holds in most normal cameras that have little or no image distortion. For example, Logitech C922 web camera has a horizontal viewing angle of 70.42° and a vertical viewing angle of 43.3° and gives nearly the same value of focal lengths with 720p image as follows.
\begin{align}
{\tilde f}_x &= \frac{1280/2}{\tan \left( 70.42\degree / 2 \right)} = 906.9217\label{eq:c922fx}\\
{\tilde f}_y &= \frac{720/2}{\tan \left( 43.4\degree / 2 \right)} = 906.9431\label{eq:c922fy}
\end{align}

However, the above conversion is hardly applicable to wide-angle or fisheye cameras. Fisheye cameras have large image distortion and the geometrical relationship in (\ref{eq:perspective}) no longer holds. For example, AXIS M2026-LE security camera has a horizontal viewing angle of 130° and a vertical viewing angle of 73°, and gives very different focal length estimates with 720p image as follows.
\begin{align}
{\tilde f}_x &= \frac{1280/2}{\tan \left( 130\degree / 2 \right)} = 298.4369\label{eq:c922fx}\\
{\tilde f}_y &= \frac{720/2}{\tan \left( 73\degree / 2 \right)} = 486.5121\label{eq:c922fy}
\end{align}

This large difference between ${\tilde f}_x$ and ${\tilde f}_y$ of wide-angle cameras obviously goes against the {\em square pixel} premise.

\subsection{Proposed calibration method}

The reason why different ${\tilde f}_x$ and ${\tilde f}_y$ are obtained for wide-angle cameras as above is that ${\tilde f}_x$ and ${\tilde f}_y$ are calculated based on the distorted image size, not the original undistorted image size.
\begin{align}
{\tilde f}_x &= {f_x}_{\mathit distorted} = \frac{W_{\mathit distorted}/2}{\tan \left( \theta_h / 2 \right)}\label{eq:fx2}\\
{\tilde f}_y &= {f_y}_{\mathit distorted} = \frac{H_{\mathit distorted}/2}{\tan \left( \theta_v / 2 \right)}\label{eq:fy2}
\end{align}

\begin{figure}[t]
\begin{center}
   \includegraphics[width=0.95\linewidth]{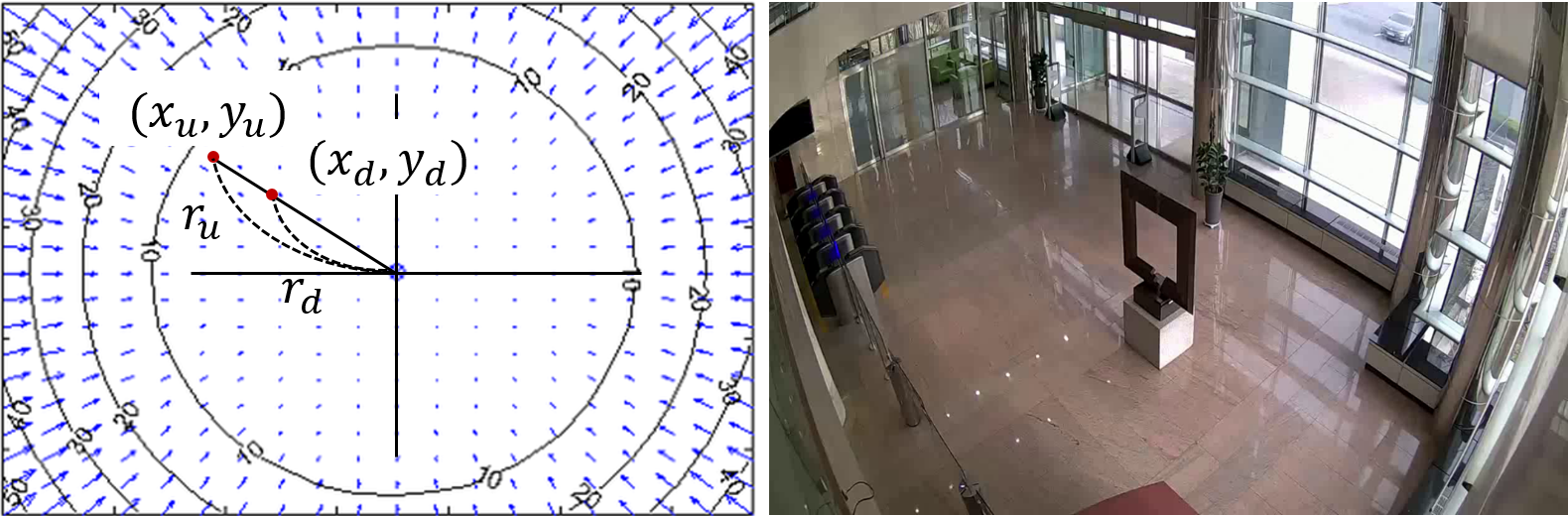}
\end{center}
   \caption{An example of radial distortion distribution~(left) and a sample image~(right).}
\label{fig:radial}
\end{figure}

The image distortion of a wide-angle camera is dominated by radial distortion. The radial distortion depends only on the distance from the principal point and the further away from the principal point, the more the image is distorted or compressed as shown in Figure~\ref{fig:radial}. Therefore, due to the characteristics of an image whose horizontal resolution is larger than vertical resolution, the more image compression occurs in the horizontal direction. As a result, ${f_x}_{\mathit distorted}$ becomes smaller than ${f_y}_{\mathit distorted}$, and this difference becomes larger as the distortion increases.

Let $\mathcal{G}$ be a mapping function that models the radial distortion of an image, $\omega$ be model parameter of $\mathcal{G}$, $r_d$ be the distance from the principal point in {\em distorted} image, and $r_u$ be the corresponding rectified distance in {\em undistorted} image.
\begin{align}
r_d &= \mathcal{G}_{\omega} (r_u)\label{eq:g}\\
r_u &= \mathcal{G}_{\omega}^{-1} (r_d)\label{eq:ginv}
\end{align}

Then, the transformation between the field of view angles and focal lengths can be reformulated as follows.
\begin{align}
{f_x}_{\mathit undistorted} &= \frac{\mathcal{G}_{\omega}^{-1}(W_{\mathit distorted}/2)}{\tan \left( \theta_h / 2 \right)}\label{eq:fxu}\\
{f_y}_{\mathit undistorted} &= \frac{\mathcal{G}_{\omega}^{-1}(H_{\mathit distorted}/2)}{\tan \left( \theta_v / 2 \right)}\label{eq:fyu}
\end{align}

And the distortion parameter $\omega$ can be estimated to minimize the difference between the computed horizontal and vertical focal lengths based on the {\em square pixel} assumption.
\begin{equation}
{\omega}^* = \arg \min_{\omega} | {f_x}_{\mathit undistorted} - {f_y}_{\mathit undistorted} |
\label{eq:w}
\end{equation}

Finally, distortion-corrected focal lengths are then obtained with the estimated $\omega^*$ as follows.
\begin{align}
f_x^* &= \frac{\mathcal{G}_{\omega^*}^{-1}(W_{\mathit distorted}/2)}{\tan \left( \theta_h / 2 \right)}\label{eq:fxu}\\
f_y^* &= \frac{\mathcal{G}_{\omega^*}^{-1}(H_{\mathit distorted}/2)}{\tan \left( \theta_v / 2 \right)}\label{eq:fyu}
\end{align}

\subsection{Modeling of radial distortion}

Radial distortion of image of wide-angle or fisheye cameras can be explained by fisheye lens projection. Unfortunately, however, there is no single fisheye lens projection, but instead fisheye lenses are usually designed to obey one of the following projections~\cite{kannala_generic_2006, panotools_fisheye_2020, wiki_fisheye_2020}:
\begin{align}
&r = 2f\tan(\theta/2) &\mbox{(stereographic)}\label{eq:stereographic}\\
&r = f\theta &\mbox{(equidistance)}\label{eq:equidistance}\\
&r = 2f\sin(\theta/2) &\mbox{(equisolid angle)}\label{eq:equisolid}\\
&r = f\sin(\theta) &\mbox{(orthographic)}\label{eq:orthographic}
\end{align}
where $\theta$ is the angle between the principal axis and the incoming ray, $r$ is the distance between the image point and the principal point and $f$ is the focal length.

Radial distortion of wide-angle cameras may be modeled by one of the above projection models. However, it is known that the equidistance projection and equisolid angle projection are most common in real fisheye lenses and stereographic and orthographic models are relatively rare~\cite{bobatkins_field_2020, panotools_fisheye_2020}. In this work, I adopt the equidistance projection model as the default distortion model considering its simplicity and popularity. However, other projection models can also be adopted depending on the specific camera lenses.

Based on equidistance projection model and from (\ref{eq:perspective}) and (\ref{eq:equidistance}), the radial distortion mapping is formulated as follows.
\begin{align}
r_d &= f\arctan \left(\frac{1}{f}r_u \right)\\
r_u &= f\tan \left(\frac{1}{f}r_d \right)
\label{eq:rdru}
\end{align}

And, by defining $\omega = 1/f$ be an adjustable distortion parameter, we can reformulate radial distortion mapping $\mathcal{G}$ for equidistance projection model as follows.
\begin{align}
\mathcal{G}_{\omega} (r_u) &= \frac{1}{\omega}\arctan \left(\omega r_u\right) \label{eq:Gw}\\
\mathcal{G}_{\omega}^{-1} (r_d) &= \frac{1}{\omega} \tan \left(\omega r_d \right) \label{eq:Gwinv}
\end{align}

\begin{figure}
\begin{center}
   \includegraphics[width=1.0\linewidth]{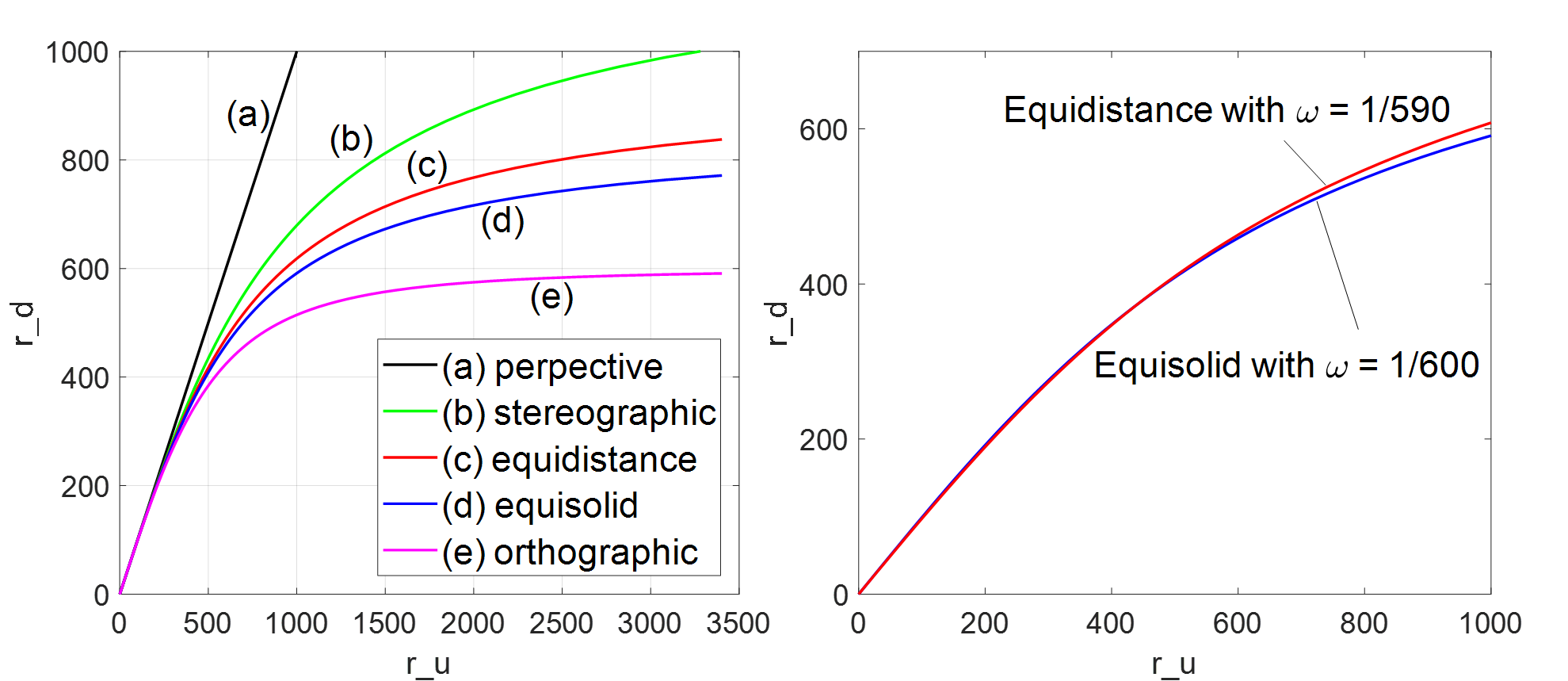}
\end{center}
   \caption{({\em Left}) Graphs of radial distortion mapping of fisheye projection models with $\omega=1/600$. ({\em Right}) Approximation of equisolid angle projection by equidistance projection. }
\label{fig:ru2rd}
\end{figure}

Figure~\ref{fig:ru2rd}({\em Left}) shows a graph of the equidistance radial distortion mapping $\mathcal{G}_{\omega}$ along with graphs of other fisheye projection models with $\omega = 1/600$. In the figure, we can see that the two common projection model ({\em equidistance} and {\em equisolid angle}) have similar mapping graph and thus they can be approximated by each other with minor error as illustrated in Figure~\ref{fig:ru2rd}({\em Right}). That means the adopted mapping function $\mathcal{G}_{\omega}$ can be applied effectively to model most fisheye and wide-angle camera lenses.

\subsection{Implementation}

With the specified distortion model $\mathcal{G}_{\omega}$, the proposed parameter estimation problem is simplified as
\begin{equation}
{\omega}^* = \arg \min_{\omega} \left| \frac{\tan(\omega W/2)}{\omega \tan(\theta_h / 2)} - \frac{\tan(\omega H/2)}{\omega \tan(\theta_v / 2)}  \right|
\label{eq:w3}
\end{equation}
\begin{equation}
f_x^* = \frac{\tan(\omega^* W/2)}{\omega^* \tan(\theta_h / 2)}
\label{eq:fxu2}
\end{equation}
\begin{equation}
f_y^* = \frac{\tan(\omega^* H/2)}{\omega^* \tan(\theta_v / 2)}
\label{eq:fyu2}
\end{equation}
where $W = W_{\mathit distorted}$ is image width and $H = H_{\mathit distorted}$ image height and $\theta_h$ is horizontal field of view angle and $\theta_v$ vertical field of view angle.

The estimation of the parameters is nearly straightforward but one caution is required. Let $J$ be the difference between the horizontal and vertical focal lengths.
\begin{equation}
J(\omega) = \frac{\tan(\omega W/2)}{\omega \tan(\theta_h / 2)} - \frac{\tan(\omega H/2)}{\omega \tan(\theta_v / 2)} \\
\end{equation}
such that
\begin{equation}
{\omega}^* = \arg \min_{\omega} \left| J(\omega) \right|.
\label{eq:w4}
\end{equation}

\begin{figure}[t]
\begin{center}
   \includegraphics[width=1.0\linewidth]{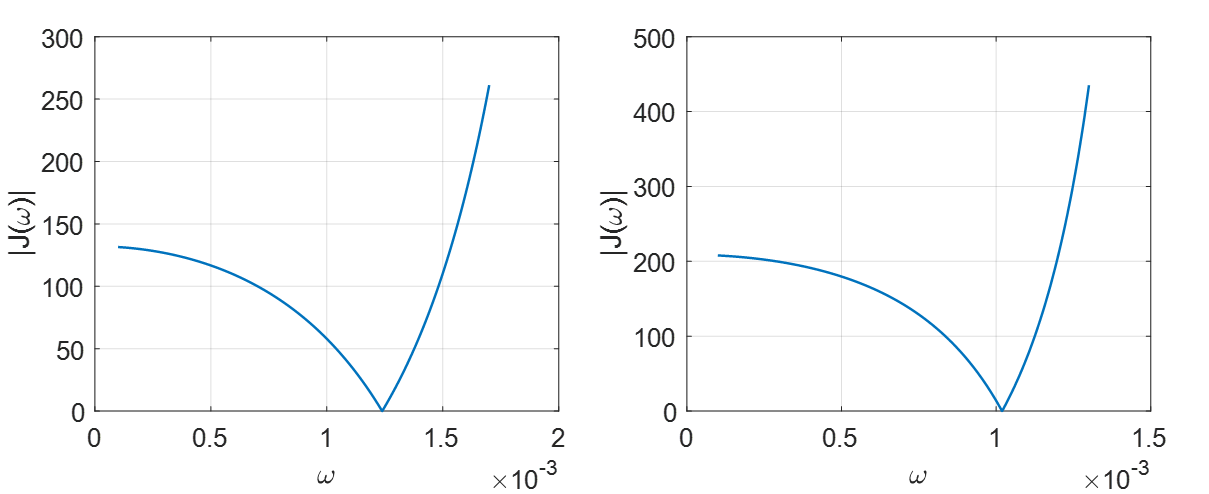}\\
\end{center}
   \caption{Graphs of $\left| J(\omega) \right|$ on two real cameras (left: L6013R, right: Hero9(W1)). }
\label{fig:Jw}
\end{figure}

As shown in Figure~\ref{fig:Jw}, the graphs of $\left| J(\omega) \right|$ plotted for real cameras show that there is a clear solution for (\ref{eq:w4}). Therefore, the distortion parameter $\omega$ would be estimated by finding a solution of $J(\omega) = 0$ by Newton–Raphson method~\cite{wallis_treatise_1685},  for $\omega \geq 0$. However, direct application of Newton-Raphson method is unstable and easy to diverge because the second derivative of $J(\omega)$ is too high~(more than 8 orders of magnitude). One solution is to reformulate $J$ so that the numerators and denominators have similar order of values. Newton-Raphson iteration with the following reformulation of $J$ always gives a stable solution and it usually converges within 30 iterations.
\begin{equation}
\omega_{k+1} = \omega_k - \gamma \frac{J_{\text{stable}}(\omega)}{\partial J_{\text{stable}}(\omega) / \partial \omega}
\end{equation}
\begin{equation}
J_{\text{stable}}(\omega) = \frac{\tan(\theta_v / 2)}{\tan(\theta_h / 2)} - \frac{\tan(\omega H/2)}{\tan(\omega W/2)}
\end{equation}
with $\omega_0 = 0.0001$ and learning rate $\gamma = 0.1$.

\section{Experiments}
\label{sec:experim}

\begin{figure*}[t]
\begin{center}
  \includegraphics[width=0.9\linewidth]{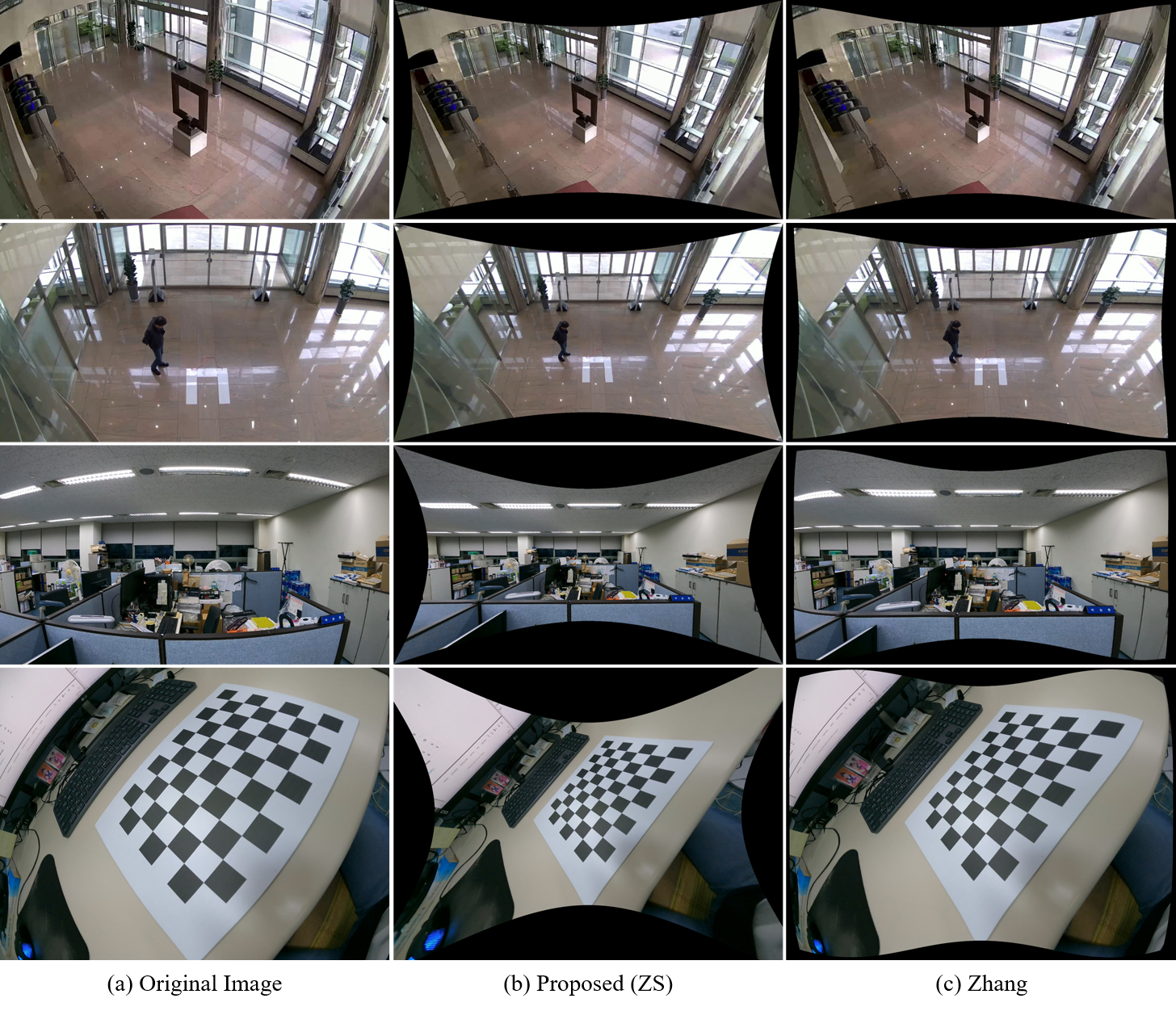}\\
\end{center}
  \vspace*{-4mm}
  \caption{Examples of rectification result of test camera images by the proposed method and Zhang's method~\cite{zhang_flexible_2000}. (b) Rectified image of (a) by proposed zero-shot calibration with camera specification only. (c) Rectified image of (a) by Zhang's method~\cite{zhang_flexible_2000} with full calibration using dozens of checkerboard images. From top to bottom: L6013R$_1$, L6013R$_2$, Hero9(W1), and Hero9(W2).}
\label{fig:undistort_comp}
\end{figure*}

\begin{table*}
\begin{center}
\begin{tabular}{ |c|rrrr|rrrr|rrr|r|  }
\hline
\multirow{2}{*}{Camera} & \multicolumn{4}{|c|}{Specification} & \multicolumn{4}{|c|}{Perspective Projection} & \multicolumn{3}{|c|}{Proposed (ZS)} & \\
& $W$ & $H$ & $\theta_h$ & $\theta_v$ & ${\tilde f}_x$ & ${\tilde f}_y$ & ${\tilde f}$ & ${\tilde f}_{\mathit err}$ & ${\omega}^*$ & $f^*$ & $f_{\mathit err}^*$ & $f_{\mathit gt}$\\
\hline
C905 & 1280 & 720 & 63.1 & - & 1042.3 & - & 1042.3 & 1.9\% & 0 & 1042.3 & 1.9\% & 1062.3\\
C922 & 1280 & 720 & 70.42 & 43.3 & 906.9 & 906.9 & 906.9 & 5.3\% & 0 & 907.0 & 5.2\% & 957.2\\
Hero9(N) & 1920 & 1080 & 73 & 45 & 1297.4 & 1303.7 & 1300.6 & 0.9\% & 0.000152 & 1306.6 & 1.4\% & 1288.8\\
Hero9(L) & 1920 & 1080 & 92 & 61 & 927.1 & 916.7 & 921.9 & 0.7\% & 0 & 921.9 & 0.7\% & 915.8\\
SNP-6321 & 1280 & 720 & 62.8 & 36.8 & 1048.5 & 1082.2 & 1065.4 & 4.2\% & 0.000570 & 1097.7 & 1.3\% & 1112.2\\
\hline
Average & & & &&&&&2.6\%&&&2.1\%&\\
\hline
\hline
L6013R$_1$ & 1280 & 720 & 86.5 & 47.8 & 680.3 & 812.4 & 746.4 & {\bf 17.7}\% & 0.001239 & 870.9 & {\bf 4.0}\% & 907.2\\
L6013R$_2$ & 1280 & 720 & 86.5 & 47.8 & 680.3 & 812.4 & 746.4 & {\bf 17.7}\% & 0.001239 & 870.9 & {\bf 4.0}\% & 906.9\\
Hero9(W1) & 1920 & 1080 & 118 & 69 & 576.8 & 785.7 & 681.3 & {\bf 22.0}\% & 0.001019 & 876.0 & {\bf 0.3}\% & 873.6\\
Hero9(W2) & 1920 & 1440 & 122 & 94 & 532.1 & 671.4 & 601.8 & {\bf 27.9}\% & 0.001051 & 837.8 & {\bf 0.4}\% & 834.5\\
M2025-LE & 1280 & 720 & 115 & 64 & 407.7 & 576.1 & 491.9 & {\bf 19.9}\% & 0.001589 & 648.8 & {\bf 5.7}\% & 613.9\\
M2026-LE & 1280 & 720 & 130 & 73 & 298.4 & 486.5 & 392.5 & {\bf 35.4}\% & 0.001775 & 568.8 & {\bf 6.3}\% & 607.1\\
\hline
Average & & & &&&&&{\bf 23.4}\%&&&{\bf 3.5}\%&\\
\hline
\end{tabular}
\end{center}
\caption{Error evaluation of the estimated focal lengths of the perspective projection and proposed zero-shot method.}
\label{table:err}
\end{table*}

In the experiment, the proposed zero-shot calibration method is evaluated qualitatively and quantitatively on 8 commercial cameras with $11$ different camera settings. And all the experiment were conducted based on the following experimental setting.

{\bf Test Cameras.} In the experiment, 8 different commercial cameras were used. The experimental settings of the test cameras and their field of view specifications are shown in Table~\ref{table:err}. The C905 and C922 are webcams, Hero9 is an action camera, and SNP-6321, L6013R$_1$, L6013R$_2$, M2025-LE, M2026-LE are surveillance cameras. The L6013R$_1$ and L6013R$_2$ are two cameras of same model. The Hero9's are physically the same camera but tested with four different digital lens settings (N: narrow, L: linear, W1, W2: wide mode). In the Table, as an exception, the vertical field of view angle of C905 is not specified because only the horizontal field of view angle is provided from the manufacturer's specification. And it should also be noted that for the M2025-LE, M2026-LE, only the specification information were used without the actual camera.

{\bf Calibration and Evaluation Dataset.} Dozens of images~(from 20 to 49) per test camera were taken carefully using a calibration checkerboard. For the M2025-LE, M2026-LE, sample images obtained from the manufacturer's Website were used instead. The captured checkerboard images were used to estimate intrinsic parameters of the test cameras and also to evaluate accuracy of the calibration methods.

{\bf Baseline Calibration Methods.} The proposed zero-shot calibration method (ZS) was compared with two standard calibration methods of Zhang~\cite{zhang_flexible_2000} and a self-implemented fisheye calibration method ($\text{FC}_\text{eq.d.}$). Zhang~\cite{zhang_flexible_2000} is the most common calibration method and is based on polynomial distortion model. The OpenCV~\cite{noauthor_opencv_2020} implementation of Zhang was used in the experiment. The second baseline method~($\text{FC}_\text{eq.d.}$) modeled image distortion by {\em equidistance} fisheye projection model described in (\ref{eq:Gw}) and (\ref{eq:Gwinv}) and can be viewed as a full calibration extension of the zero-shot method. I implemented $\text{FC}_\text{eq.d.}$ so that it estimates parameters so as to minimize the sum of pixel reprojection errors by using the Levenberg-Marquardt algorithm~\cite{more_levenberg-marquardt_1978}.

\subsection{Qualitative evaluation}

In the first experiment, I evaluate the quality of the camera parameters estimated with the proposed zero-shot calibration method. The estimated distortion parameters~($\omega^*$) and focal lengths~($f^*$) for the 11 test cameras are shown in Table~\ref{table:err}.

{\bf Quality of Distortion Parameter.} The quality of the estimated distortion parameters is evaluated by comparing the rectification results of the test camera images.

Figure~\ref{fig:undistort_comp} shows rectification result of sample images of test cameras by the proposed method and Zhang's method~\cite{zhang_flexible_2000}. As we can see that, the images are surprisingly well rectified by the proposed method although only the camera specifications were used. And also, the results are shown to be better than those of Zhang's method~\cite{zhang_flexible_2000}, where the camera parameters were estimated by full calibration using dozens of the checkerboard images.

\begin{figure}[t]
\begin{center}
  \includegraphics[width=0.97\linewidth]{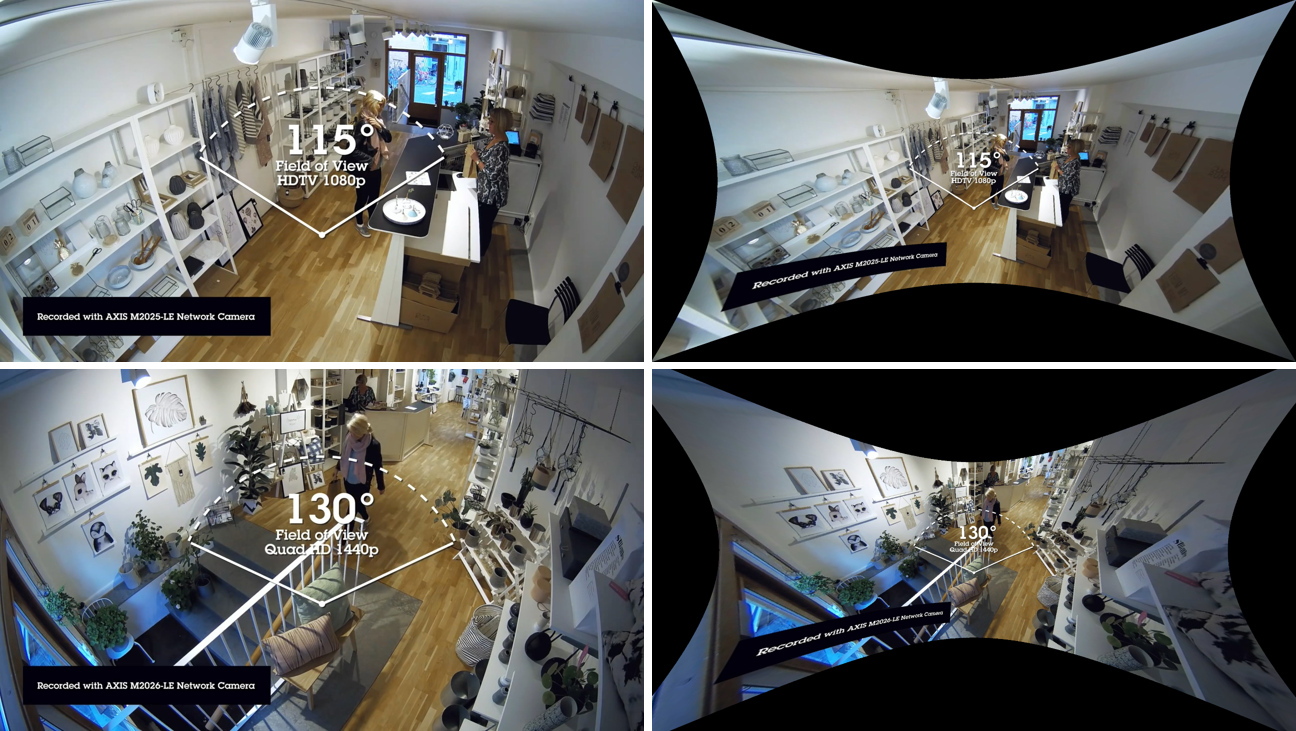}
\end{center}
  \caption{Sample rectification results of internet images with their camera specification by proposed zero-shot calibration method. The cameras are M2025-LE and M2026-LE from top to bottom.}
\label{fig:undistort}
\end{figure}

As another example, Figure~\ref{fig:undistort} shows sample rectification results of internet images with their camera specification by the proposed zero-shot calibration method. The cameras are those that I don't really have, and thus test sample images were obtained from the manufacturer's Website. The results also show that the images are well rectified up to image corners.

{\bf Quality of Focal Length.} The estimated focal lengths are evaluated by the following deviation error measure.
\begin{equation}
f_{err} = \frac{|f - f_{gt}|}{f_{gt}}
\label{eq:ferr}
\end{equation}
where $f$ is the estimate focal length and $f_{gt}$ is ground truth value. Since actual ground truth is unknown, focal lengths obtained from full calibration of Zhang's method~\cite{zhang_flexible_2000} were used as a ground truth instead. In cases of M2025-LE and M2026-LE, only the specification information were used without real cameras and their ground truth focal lengths were manually estimated with internet image samples using the method of Košecká and Zhang~\cite{kosecka_video_2002}.

The evaluated focal length errors for perspective projection and proposed method are shown in Table~\ref{table:err}, where the cameras are sorted in two groups so that the narrow-angle cameras are listed at the top and wide-angle cameras listed down. First, the estimated focal lengths by perspective projection are shown by ${\tilde f}_x$ and ${\tilde f}_y$ in Table~\ref{table:err}. As previously described, they show large difference for wide-angle cameras while similar for narrow-angle cameras. And their averaged value ${\tilde f}$ are shown to give large deviation error from the ground truth of an average of 23.4\%. We can see that the deviation error for perspective projection increases roughly proportional to the field of view angles of the camera. The high deviation error of perspective projection is because it does not reflect the image distortion of a camera. On the other hand, the proposed zero-shot method estimates distortion parameters ($\omega$) directly from the camera specification and then estimates distortion-corrected focal lengths, giving more accurate estimate. The results in Table~\ref{table:err} show that by applying the proposed zero-shot method the deviation errors were greatly reduced from 23.4\% to 3.5\% on average for wide-angle cameras. As for narrow-angle cameras, the error also slightly decreased from 2.6\% to 2.1\%.

\subsection{Quantitative evaluation}

In the second experiment, I evaluate accuracy of the proposed method in terms of pixel reprojection error.

Let $N$ calibration images was obtained for a camera and let each calibration image have $M$ control points. And let $p_{ij}$ be the measured $j$-th control point of $i$-the calibration image and ${\tilde p}_{ij} = K[R_i|t_i]P_{ij}$ be the projected control point using the estimated camera parameters of the camera, where $K$ is estimated camera parameter, $R_i$ and $t_i$ are the external camera parameters of $i$-th calibration image, and $P_{ij}$ is the world coordinate of the control point. The external camera parameters were estimated for each calibration image from the correspondences between $K^{-1}p_{ij}$ and $P_{ij}$ using the PnP algorithm~\cite{lepetit_epnp_2008}. Then the root mean square~(RMS) pixel reprojection error $\epsilon$ is computed as follows.
\begin{equation}
\epsilon^2 = \frac{1}{NM} \sum_{i=1}^{N}\sum_{j=1}^{M}{\| p_{ij} - {\tilde p}_{ij}\|^2}
\label{eq:rms}
\end{equation}

In the experiment, I first estimated camera parameters of 9 cameras from their specifications using the proposed method, and then computed the RMS reprojection errors using the estimated camera parameters on the test images set. The result is shown in Table~\ref{table:rms0}.

\begin{table}
\begin{center}
\begin{tabular}{ |c|r|rr|rr|  }
\hline
 & \multirow{2}{*}{N} & Pers. & ZS & Zhang & $\text{FC}_\text{eq.d.}$ \\
 & & $\epsilon$& $\epsilon$& $\epsilon$& $\epsilon$\\
\hline
C905 & 20 & 0.98 & 0.98 & 0.49 & 0.71 \\
C922 & 22 & 1.18 & 1.18 & 0.32 & 0.51\\
Hero9(N) & 22 & 0.78 & 0.91  & 0.53& 0.57\\
Hero9(L) & 25 & 1.00 & 1.00& 0.43 & 0.50 \\
SNP-6321 & 30 & 2.59 & 0.69  & 0.37& 0.49	\\
\hline
Average & 23.8 & 1.31 & 0.95 & 0.43& 0.55 \\
\hline
\hline
L6013R$_1$ & 49 & 15.50 & 2.15 & 0.69& 1.03 \\
L6013R$_2$ & 23 & 14.32 & 1.50 & 0.40& 0.51 \\
Hero9(W1) & 30 & 37.78 & 0.48 & 1.04& 0.47 \\
Hero9(W2) & 29 & 45.01 & 0.76 & 1.16& 0.75 \\
\hline
Average & 32.8 & 28.15 & 1.22 & 0.82& 0.69 \\
\hline
\end{tabular}
\end{center}
\caption{The RMS reprojection error in pixels.}
\label{table:rms0}
\end{table}

The proposed approach is shown to give an average of 1.22 RMS error for wide-angle cameras despite using only camera specification information. And this result is quite comparable to Zhang~\cite{zhang_flexible_2000}, which had performed full optimization on the evaluation images and gave 0.82 RMS error on average. As for narrow-angle cameras, the proposed method (ZS) gave 0.95 RMS error and Zhang gave 0.43 RMS error on average.

For the reference, results using other calibration methods are also presented in Table~\ref{table:rms0}. As expected, the simple perspective projection estimation~(Pers.) gave the most high RMS error of 28.15 on average for wide-angle cameras but an average of 1.31 RMS error for narrow-angle cameras. As for fisheye calibration method~($\text{FC}_\text{eq.d.}$), it gave an average of 0.69 RMS error for wide-angle cameras, which is better than Zhang. However, Zhang is shown to give better result than $\text{FC}_\text{eq.d.}$ for narrow-angle cameras.

\section{Application to Single-Shot Calibration}
\label{sec:singleshot}

\begin{table*}
\begin{center}
\begin{tabular}{ |c|c| r@{~}l | r@{~}l | r| r@{~}l|r|}
\hline
 & \multirow{2}{*}{N} & \multicolumn{2}{|c|}{$\text{FC}_\text{eq.d.}^{(1)}$} & \multicolumn{2}{|c|}{Zhang$^{(1)}$} & ZS & \multicolumn{2}{|c|}{$\text{FC}_\text{eq.d.}^{(1)}$ + ZS} & Zhang$^{(\text{N})}$ \\
 &&${\bar \epsilon}_{gen}$&(${\bar \epsilon}_{fit}$)&${\bar \epsilon}_{gen}$&(${\bar \epsilon}_{fit}$)&$\epsilon$&${\bar \epsilon}_{gen}$&(${\bar \epsilon}_{fit})$&$\epsilon$\\
\hline
C905 & 20 & 6.48&(5.35)&1.06&(0.35)&0.98&~~~~~~0.89&(0.64)&0.49\\
C922 & 22 & 3.49&(3.03)&0.52&(0.17)&1.18&0.98&(0.84)&0.32\\
Hero9(N) & 22 & 3.67&(3.34)&0.77&(0.46)&0.91&0.68&(0.59)&0.53\\
Hero9(L) & 25 & 5.80&(5.43)&0.67&(0.37)&1.00&0.97&(0.74)&0.43\\
SNP-6321 & 30 & 4.96&(3.80)&1.08&(0.23)&0.69&0.59&(0.41)&0.37\\
\hline
Average & 23.8 & 4.88&(4.19)&0.82&(0.32)&0.95&0.82&(0.64)&0.43\\
\hline
\hline
L6013R$_1$ & 49 & {\bf 4.20} &(2.20)&0.90&(0.26)&1.50&{\bf 0.88}&(0.68)&0.69\\
L6013R$_2$ & 23 & {\bf 6.26} &(2.65)&2.21&(0.39)&2.15&{\bf 1.54}&(1.13)&0.40\\
Hero9(W1) & 30 & {\bf 10.87} &(7.36)&3.41&(0.56)&0.48&{\bf 0.53}&(0.44)&1.04\\
Hero9(W2) & 29 & {\bf 15.55} &(13.95)&2.22&(0.69)&0.76&{\bf 0.81}&(0.70)&1.16\\
\hline
Average & 32.8 & {\bf 9.22} &(6.54)&2.19&(0.48)&1.22&{\bf 0.94}&(0.73)&0.82\\
\hline
\end{tabular}
\end{center}
\caption{Evaluation result of single-shot calibration. N denotes the number of images obtained for each test camera.}
\label{table:rms1}
\end{table*}

The primary purpose of the proposed method is to calibrate cameras without any image acquisition nor image processing. However, it can easily be extended as a single-shot or few-shot approach. The method to extend is to first find the initial parameters with the zero-shot method and then refine them on calibration images. The estimated initial parameters can also be used to provide proper initialization of any existing calibration methods.

An additional experiment was conducted to evaluate the effectiveness of this single-shot application. Here, the calibration is performed with only one calibration image. And then the estimated parameters are evaluated over all calibration images, giving a generalization error. In order to evaluate single-shot calibration methods properly, I defined two error metric. One is fitting RMS error and the other is generalization RMS error. For a given single-shot calibration method and a test camera with $N$ calibration images, an {\em average generalization error}~(${\bar \epsilon}_{gen}$) and {\em average fitting error}~(${\bar \epsilon}_{fit}$) are computed as follows.
\begin{align}
{\bar \epsilon}_{gen} &= \frac{1}{N} \sum_{k=1}^{N} \sqrt{\frac{1}{NM} \sum_{i=1}^{N}\sum_{j=1}^{M}{\| p_{ij} - {\tilde p}_{ij}^{(k)}\|^2}}\label{eq:rmsgen}\\
{\bar \epsilon}_{fit} &= \frac{1}{N} \sum_{k=1}^{N} \sqrt{\frac{1}{M} \sum_{j=1}^{M}{\| p_{ij} - {\tilde p}_{ij}^{(k)}\|^2}}\label{eq:rmsfit}
\end{align}
where ${\tilde p}_{ij}^{(k)}$ is the projected control point using the single-shot calibration result which have calibrated on $k$-th calibration image only. Note that the {\em single-shot} calibration is performed for each image and then the evaluation results are averaged.

The evaluation results are shown in Table~\ref{table:rms1}. First, the single-shot fisheye calibration~($\text{FC}_\text{eq.d.}^{(1)}$) is shown to give 9.22 RMS generalization error and 6.54 RMS fitting error on average for wide-angle cameras~(the superscripted numbers in the method name denote the number of images used for calibration). Zhang$^{(1)}$ gave better result of 2.19 RMS generalization error with 0.48 RMS fitting error for wide-angle cameras on average but it is still worse than the zero-shot result of 1.22 RMS error.

Those results imply that the single-shot calibration methods are easy to fail or overfit the image, giving high generalization error. However, if we combine single-shot calibration with the proposed zero-shot calibration~(ZS) such that ZS provides initial estimates of the parameters, we can expect that the methods converge more stably or avoid local minima. The experimental result in Table~\ref{table:rms1} shows that with the combination of $\text{FC}_\text{eq.d.}^{(1)}$ and ZS, the RMS generalization error was greatly reduced from 9.22 to 0.94 for wide-angle cameras, justifying the effectiveness of this application. The combined calibration method is still a single-shot method but its performance was comparable to Zhang's full calibration result (Zhang$^{(\text{N})}$) which gave 0.82 RMS error.

\section{Discussion and Conclusion}
\label{sec:discussion}

In this paper, I presented a novel zero-shot calibration method that estimates camera parameters with no calibration image. The experimental results on eight commercial cameras showed that the proposed method was able to estimate their camera parameters effectively with an accuracy comparable to conventional image-based calibration method. Moreover, the estimated camera parameters by the method can also be used to provide proper initialization of any existing calibration methods, and the effectiveness of this kind of initialization was also demonstrated in the experiment.

{\bf Applications.} My work is the first one to estimate camera parameters of fisheye cameras without image acquisition and it is expected to be practically utilized in various applications. For example, it can be used as a practical alternative in field applications where individual calibration is difficult or infeasible, and in most visual surveillance applications where calibration accuracy is less critical.

{\bf Adaptive Selection of Fisheye Model.} In this paper, I adopted an equidistance fisheye projection model as our default distortion model. It is expected to work for most cases but it can be sometime problematic when a camera deviates from the model. One solution is to select distortion model adaptively depending on the cameras. The suggested zero-shot method can be extended to select distortion model adaptively if a diagonal field of view angle is additionally available. The horizontal, vertical, and diagonal view angles give three constraints on the radial mapping and then a model that best meets these constraints can be selected.

{\bf Limitations.} My approach has several limitations. First, it cannot be applied for the cameras that their specification are not provided. And the cameras that have wide-range varifocal lenses are also not applicable. The propose method is particularly useful for fisheye cameras that have large image distortion but it's not very useful for narrow-angle cameras. However, even a narrow-angle camera, it can be effectively utilized when the camera has image distortion.

{\bf Future Works.} One future work is to evaluate the influence of camera auto-focusing and the accuracy of the field of view angles provided by the manufacturers. Another work is to combine the method with more conventional calibration methods and evaluate their effectiveness. Quantitative comparison with recent deep learning-based approaches on benchmark data set is also another work. Last one is to compare the influence of fisheye projection model selection on the performance on real cameras and to extend the method to select distortion model adaptively.

\section*{Acknowledgement}

This work was supported by the ICT R\&D program of MSIT/IITP. [2019-0-01309, Development of AI Technology for Guidance of a Mobile Robot to its Goal with Uncertain Maps in Indoor/Outdoor Environments]

{\small
\bibliographystyle{ieee_fullname}
\bibliography{zeroshot_cvpr21}

\begin{thebibliography}{10}\itemsep=-1pt

\bibitem{noauthor_opencv_2020}
{OpenCV}.
\newblock \url{https://opencv.org/}.

\bibitem{bobatkins_field_2020}
B. Akins.
\newblock Field of {View} - {Rectilinear} and {Fisheye} lenses.
\newblock
  \url{http://www.bobatkins.com/photography/technical/field\_of\_view.html}.

\bibitem{beardsley92}
Paul Beardsley and David Murray.
\newblock Camera calibration using vanishing points.
\newblock In {\em BMVC92}, pages 416--425. Springer, 1992.

\bibitem{chao_self-supervised_2020}
C. Chao, P. Hsu, H. Lee, and Y.~F. Wang.
\newblock Self-{Supervised} {Deep} {Learning} for {Fisheye} {Image}
  {Rectification}.
\newblock In {\em {ICASSP} 2020 - 2020 {IEEE} {International} {Conference} on
  {Acoustics}, {Speech} and {Signal} {Processing} ({ICASSP})}, pages
  2248--2252, May 2020.
\newblock ISSN: 2379-190X.

\bibitem{zhou_chuan_planar_2003}
Zhou Chuan, Tan~Da Long, Zhu Feng, and Dong~Zai Li.
\newblock A planar homography estimation method for camera calibration.
\newblock In {\em Proceedings 2003 {IEEE} {International} {Symposium} on
  {Computational} {Intelligence} in {Robotics} and {Automation}}, volume~1,
  pages 424--429, July 2003.

\bibitem{galeone_camera_2018}
Paolo Galeone.
\newblock Camera calibration guidelines, 2018.
\newblock
  https://pgaleone.eu/computer-vision/2018/03/04/camera-calibration-guidelines/.

\bibitem{grossmann_camera_2014}
Etienne Grossmann, John~Iselin Woodfill, and Gaile~Gibson Gordon.
\newblock Camera calibration using an easily produced {3D} calibration pattern,
  Oct. 2014.

\bibitem{guillou00}
Erwan Guillou, Daniel Meneveaux, Eric Maisel, and Kadi Bouatouch.
\newblock Using vanishing points for camera calibration and coarse 3d
  reconstruction from a single image.
\newblock {\em The Visual Computer}, 16(7):396--410, 2000.

\bibitem{herrera_forget_2016}
Daniel Herrera, C.~Juho Kannala, and Janne Heikkila.
\newblock Forget the checkerboard: {Practical} self-calibration using a planar
  scene.
\newblock In {\em 2016 {IEEE} {Winter} {Conference} on {Applications} of
  {Computer} {Vision} ({WACV})}, pages 1--9, Lake Placid, NY, USA, Mar. 2016.
  IEEE.

\bibitem{huang_research_2019}
Lin Huang, Feipeng Da, and Shaoyan Gai.
\newblock Research on multi-camera calibration and point cloud correction
  method based on three-dimensional calibration object.
\newblock {\em Optics and Lasers in Engineering}, 115:32--41, Apr. 2019.

\bibitem{kannala_generic_2006}
J. Kannala and S.S. Brandt.
\newblock A generic camera model and calibration method for conventional,
  wide-angle, and fish-eye lenses.
\newblock {\em IEEE Transactions on Pattern Analysis and Machine Intelligence},
  28(8):1335--1340, Aug. 2006.

\bibitem{kosecka_video_2002}
Jana Košecká and Wei Zhang.
\newblock Video {Compass}.
\newblock In {\em European conference on computer vision}, Lecture {Notes} in
  {Computer} {Science}, pages 476--490, Berlin, Heidelberg, 2002. Springer.

\bibitem{lee2012camera}
Joo-Haeng Lee.
\newblock Camera calibration from a single image based on coupled line cameras
  and rectangle constraint.
\newblock In {\em Pattern Recognition (ICPR), 2012 21st International
  Conference on}, pages 758--762. IEEE, 2012.

\bibitem{lepetit_epnp_2008}
Vincent Lepetit, Francesc Moreno-Noguer, and Pascal Fua.
\newblock {EPnP}: {An} {Accurate} {O}(n) {Solution} to the {PnP} {Problem}.
\newblock {\em International Journal of Computer Vision}, 81(2):155, July 2008.

\bibitem{xiaoyu_blind_2019}
Xiaoyu Li, Bo Zhang, Pedro~V. Sander, and Jing Liao.
\newblock Blind geometric distortion correction on images through deep
  learning.
\newblock In {\em IEEE Conf. Comput. Vis. Pattern Recog.}, May 2019.

\bibitem{liao_deep_2020}
Kang Liao, Chunyu Lin, and Yao Zhao.
\newblock A {Deep} {Ordinal} {Distortion} {Estimation} {Approach} for
  {Distortion} {Rectification}.
\newblock {\em arXiv:2007.10689 [cs]}, July 2020.
\newblock arXiv: 2007.10689.

\bibitem{lv_methods_nodate}
Yaowen Lv, Wei Liu, and Xiping Xu.
\newblock Methods based on {1D} homography for camera calibration with {1D}
  objects.
\newblock {\em Applied Optics}, 57(9):2155--2164, Mar. 2018.

\bibitem{more_levenberg-marquardt_1978}
Jorge~J. Moré.
\newblock {\em The {Levenberg}-{Marquardt} algorithm: {Implementation} and
  theory}.
\newblock Lecture {Notes} in {Mathematics}. Springer, Berlin, Heidelberg, 1978.

\bibitem{panotools_fisheye_2020}
PanoTools.
\newblock Fisheye {Projection}.
\newblock \url{https://wiki.panotools.org/Fisheye\_Projection}, 2020.

\bibitem{strobl_more_2011}
Klaus~H. Strobl and Gerd Hirzinger.
\newblock More accurate pinhole camera calibration with imperfect planar
  target.
\newblock In {\em 2011 {IEEE} {International} {Conference} on {Computer}
  {Vision} {Workshops} ({ICCV} {Workshops})}, pages 1068--1075, Barcelona,
  Spain, Nov. 2011. IEEE.

\bibitem{wallis_treatise_1685}
John Wallis.
\newblock {\em A {Treatise} of {Algebra}, {Both} {Historical} and {Practica}}.
\newblock Oxford: Richard Davis, 1685.

\bibitem{wiki_fisheye_2020}
Wikipedia.
\newblock Fisheye lens.
\newblock
  \url{https://en.wikipedia.org/w/index.php?title=Fisheye\_lens\&oldid=986382882}.

\bibitem{xue_learning_2019}
Zhucun Xue, Nan Xue, Gui-Song Xia, and Weiming Shen.
\newblock Learning to {Calibrate} {Straight} {Lines} for {Fisheye} {Image}
  {Rectification}.
\newblock In {\em IEEE Conf. Comput. Vis. Pattern Recog.}, May 2019.

\bibitem{xue_fisheye_2020}
Zhu-Cun Xue, Nan Xue, and Gui-Song Xia.
\newblock Fisheye {Distortion} {Rectification} from {Deep} {Straight} {Lines}.
\newblock {\em arXiv:2003.11386 [cs]}, Mar. 2020.
\newblock arXiv: 2003.11386.

\bibitem{xiaoqing_fisheye_2018}
Xiaoqing Yin, Xinchao Wang, Jun Yu, Maojun Zhang, Pascal Fua, and Dacheng Tao.
\newblock Fisheyerecnet: A multi-context collaborative deep network for fisheye
  image rectification.
\newblock In {\em Eur. Conf. Comput. Vis.}, May 2018.

\bibitem{zhang_homography-based_2008}
Beiwei Zhang and Youfu Li.
\newblock Homography-based method for calibrating an omnidirectional vision
  system.
\newblock {\em Journal of the Optical Society of America A}, 25(6):1389--1394,
  June 2008.

\bibitem{zhang_flexible_2000}
Zhengyou Zhang.
\newblock A flexible new technique for camera calibration.
\newblock {\em IEEE Transactions on pattern analysis and machine intelligence},
  22(11):1330--1334, Nov. 2000.

\bibitem{zhao_simple_2020}
H. Zhao, Y. Shi, X. Tong, X. Ying, and H. Zha.
\newblock A {Simple} {Yet} {Effective} {Pipeline} {For} {Radial} {Distortion}
  {Correction}.
\newblock In {\em 2020 {IEEE} {International} {Conference} on {Image}
  {Processing} ({ICIP})}, pages 878--882, Oct. 2020.
\newblock ISSN: 2381-8549.

\bibitem{zhengyou_zhang_flexible_1999}
{Zhengyou Zhang}.
\newblock Flexible camera calibration by viewing a plane from unknown
  orientations.
\newblock In {\em Proceedings of the {Seventh} {IEEE} {International}
  {Conference} on {Computer} {Vision}}, pages 666--673 vol.1, Kerkyra, Greece,
  1999. IEEE.

\end{thebibliography}
}

\end{document}